\documentclass{article}

\usepackage{arxiv}
\usepackage[utf8]{inputenc}
\usepackage[T1]{fontenc}
\usepackage{url}
\usepackage{booktabs}
\usepackage{amsfonts} 
\usepackage{nicefrac}
\usepackage{microtype}
\usepackage{lipsum}
\usepackage{graphicx}
\usepackage{cite}
\usepackage{doi}
\usepackage{microtype}
\usepackage{subfigure}
\usepackage{booktabs}
\usepackage{amsfonts}
\usepackage{amssymb}
\usepackage[normalem]{ulem}
\usepackage[dvipsnames]{xcolor}
\usepackage{xspace}
\usepackage{listings}
\usepackage{inconsolata}
\usepackage{mathtools}
\usepackage{hyperref}
\usepackage[noabbrev, capitalise, nameinlink]{cleveref}
\usepackage[flushleft]{threeparttable}
\usepackage{multirow}
\usepackage{makecell}
\newcommand{\tablefontsize}{\small}
\usepackage{enumitem}
\newtheorem{theorem}{Theorem}

\pdfoutput=1

\newcommand{\idea}{KAS\xspace}
\newcommand{\ideafull}{Kernel Architecture Search\xspace}
\newcommand{\name}{Canvas\xspace}
\newcommand{\udag}{micro-DAG\xspace}
\newcommand{\papertitle}{\name: End-to-End \ideafull in Neural Networks}

\newcommand{\codeword}[1]{\texttt{\textcolor{black}{#1}}}

\newcommand{\tensorshape}[1]{$[#1]$}

\newcommand{\delete}[1]{}

\title{\papertitle}
\date{}

\author{Chenggang Zhao \\ 
	Institute for Interdisciplinary Information Sciences \\
	Tsinghua University \\
	Beijing, China \\
	\texttt{zhaocg21@mails.tsinghua.edu.cn} \\
	\And
	Genghan Zhang \\
	Department of Electrical Engineering \\
        Tsinghua University \\
	Beijing, China \\
	\texttt{ghzhang19@gmail.com} \\
        \And
	Ao Shen \\
	Department of Computer and Information Technology \\
        Purdue University \\
	West Lafayette, IN 47907, USA \\
	\texttt{shen634@purdue.edu} \\
        \And
	  Mingyu Gao \\
	Institute for Interdisciplinary Information Sciences \\
        Tsinghua University \\
	Beijing, China \\
	\texttt{gaomy@tsinghua.edu.cn} \\
}

\renewcommand{\headeright}{}
\renewcommand{\undertitle}{}

\hypersetup{
pdftitle={\papertitle},
pdfauthor={Chenggang Zhao, Genghan Zhang, Mingyu Gao},
pdfkeywords={Machine Learning, Tensor Compilers, Neural Architecture Search},
}

\begin{document}
\maketitle

\vspace{-15pt}

\begin{abstract}
The demands for higher performance and accuracy in neural networks (NNs) never end.
Existing tensor compilation and Neural Architecture Search (NAS) techniques orthogonally optimize the two goals but actually share many similarities in their concrete strategies.
We exploit such opportunities by combining the two into one and make a case for \ideafull (\idea).
\idea reviews NAS from a system perspective and zooms into a more fine-grained level to generate neural kernels with both high performance and good accuracy.
To demonstrate the potential of \idea, we build an end-to-end framework, \name, to find high-quality kernels as convolution replacements.
\name samples from a rich set of fine-grained primitives to stochastically and iteratively construct new kernels and evaluate them according to user-specified constraints.
\name supports freely adjustable tensor dimension sizes inside the kernel and uses two levels of solvers to satisfy structural legality and fully utilize model budgets.
The evaluation shows that by replacing standard convolutions with generated new kernels in common NNs, \name achieves average $1.5\times$ speedups compared to the previous state-of-the-art with acceptable accuracy loss and search efficiency.
\name verifies the practicability of \idea by rediscovering many manually designed kernels in the past and producing new structures that may inspire future machine learning innovations.
For source code and implementation, we open-sourced \name at \href{https://github.com/tsinghua-ideal/Canvas}{https://github.com/tsinghua-ideal/Canvas}.

\end{abstract}

\keywords{Machine Learning \and Tensor Compilers \and Neural Architecture Search}

\section{Introduction}

Many emerging techniques backed by neural networks (NNs), including computer vision, natural language processing, and robotics, have boomed these years and made tremendous progress.
However, NNs remain as a complex algorithm that not only consumes significant computation resources to achieve acceptable (e.g., real-time) performance but also lacks an effective and theoretical methodology to design models with high accuracy.
The demands for high performance and high accuracy only keep increasing when NNs are applied to more scenarios in the real world. 

System and machine learning communities have adopted orthogonal approaches to satisfy the high demands above.
From a system perspective, NNs are represented as tensor programs, and specialized tensor compilers and frameworks have been developed to realize high-performance execution on different hardware platforms.
On the other hand, NN algorithm researchers have started to use automatic methods to design better model architectures with improved accuracy, known as Neural Architecture Search (NAS).
Despite the different goals, these two directions share many common underlying techniques.
Both attempt to reorganize the NN structures and redistribute the computations by transforming their basic blocks in the rich design spaces; both require timely evaluation as the feedback to guide the exploration.
Consequently, recent efforts have started to exploit these similarities and simultaneously conduct performance and accuracy exploration with a careful tradeoff between the two. Examples include giving up mathematical equivalence in tensor compilers to unleash higher performance~\cite{wang2021pet, turner}, and making NAS aware of performance besides the accuracy goal~\cite{cai2018proxylessnas, tan2019mnasnet, tan2019efficientnet, wang2019haq}.

In this paper, we take these opportunities one step further and make a case for a new paradigm of \emph{\ideafull} (\idea).
To maximize runtime performance while pursuing high model accuracy, \idea searches for an efficient \emph{kernel architecture} from a 
lower-level system perspective.
It then uses the generated kernels to replace the standard layers (e.g., convolutions) in existing NNs.
\idea \emph{stochastically and iteratively} builds candidate kernels using a rich set of \emph{fine-grained primitives}.
It treats \emph{runtime performance as first-priority constraints} and searches for kernels that achieve the best accuracy within the given performance limit.
By completely discarding mathematical equivalence and searching for new designs, \idea enables higher performance than tensor compilers.
By primarily focusing on runtime and composing kernels in a fine-grained way, \idea complements NAS from a system perspective while retaining similar levels of accuracy.

To demonstrate the promising potentials of \idea, we further build an end-to-end, automated, and efficient framework, \name, that could find new high-quality kernels to replace traditional convolutions.
\name relies on a \textbf{fine-grained primitive library} as the building blocks for kernel construction, following the philosophy of \emph{decoupling data rearrangements and arithmetic computations},
The primitive designs are inspired by the decoupling concept in system research while at the same time also having great expressibility to realize complicated mathematical compositions in machine learning.
From such a primitive library, \name uses a \textbf{random sampler} to \emph{stochastically and iteratively generate candidate kernels} from a large search space.
The sampler carefully adjusts the sampling probabilities of different primitive types to ensure fair and legitimate kernel construction. It also applies various pruning rules.
Then \name evaluates the candidate kernels using two classes of \textbf{user-specified constraints and optimization goals}, including those that can be \emph{analytically} modeled and calculated (e.g., numbers of operations and parameters), and those that need \emph{experimental} measurements (e.g., runtime and accuracy).
Finally, a key innovation of \name is the use of free \emph{dynamic variables} on tensor dimension sizes during kernel generation.
Dynamic variables greatly enrich the search space but must eventually be substituted with reasonable values that satisfy all structural legality and fully utilize allowed model budgets.
So we design \textbf{two levels of dynamic variable solvers} in \name to handle the two requirements with provable correctness and high efficiency.

On top of several popular backbone NNs, we use \name to search for the best kernels to replace standard convolutions.
Compared with the original model optimized by TVM Ansor~\cite{zheng2020ansor}, \name achieves on average $1.5\times$ and up to $4.6\times$ speedups with acceptable accuracy loss ($\sim$ 1\%).
And models are reduced to $0.1 \sim 0.6\times$ of the original sizes.
\name also outperforms a previous work that searches for efficient kernel implementations in a network-independent way and then applies manual layer-wise optimizations~\cite{turner}, by $1.4 \sim 1.6\times$ faster.
We also conduct detailed case studies to examine what kinds of kernels \name discovers.
Interestingly, \name rediscovers many manually proposed kernel designs in the past and produces previously unexplored structures that may inspire future NN design automation and innovations.
\section{Background}
\label{sec:background}

To continuously improve the efficiency and effectiveness of neural networks (NNs), the two research communities of computer systems and machine learning have taken different but complementary approaches in the past years.
From the system perspective, there is a large design space about how to carry out the computations specified by the mathematical representation of an NN, resulting in potential orders of magnitude runtime performance differences on real hardware platforms like CPUs, GPUs, and specialized chips.
To ease the efforts of programming and optimization, many frameworks, including PyTorch~\cite{paszke2019pytorch}, TensorFlow~\cite{abadi2016tensorflow}, and TVM~\cite{chen2018tvm, autotvm}, have been proposed.
They typically represent NNs as \emph{tensor programs} and apply fine-grained optimizations at multiple levels, from the computation graph~\cite{wang2021pet, taso, yang2021equality, ding2021ios} to the loop nest of an individual operator kernel~\cite{chen2018tvm, zheng2020ansor}.

Meanwhile, \emph{Neural Architecture Search} (NAS) has become an increasingly popular methodology for designing effective NN architectures.
Indeed, a large part of the recently proposed state-of-the-art NNs is automatically discovered by NAS rather than composed manually~\cite{tan2019mnasnet, zoph2018learning, liu2018darts, liu2018progressive, wu2019fbnet}.
NAS typically defines a highly modular search space by dividing the backbone network topology into basic units or cells of various sizes.
Then it searches for how to build each cell by connecting basic layers like convolution and pooling.
During the search, the accuracy levels of the candidate cell structures are continuously evaluated using statistic metrics or through training sample datasets.
In some sense, NAS also organizes NNs towards a better evaluation objective, but rather on model accuracy than runtime performance.

Regardless of the research perspectives, the essence of the problem lies in getting reliable accuracy levels with a demand for less computation.
It is only the overall research context that leads to the two communities of systems and machine learning focusing on different approaches: the system side leverages fine-grained performance optimizations under the rigid constraint of mathematical equivalence; in contrast, the ML side is free to transform the mathematical form of the NN at a coarse-grained level (i.e., layer-wise operators) to improve accuracy.

Turner et al.~\cite{turner} first captured this new opportunity.
Specifically, they introduced two NAS-style transformations into the current TVM~\cite{chen2018tvm} compilation framework: bottlenecking (reducing channel counts) and grouping (dividing channels into groups). The new transformations gave up traditional compiler transformation equivalence. 
However, this work was still preliminary, as it only changed the loop ranges while retaining the original loop structure of traditional convolutions, leaving abundant opportunities unexplored.
Also, the workflow was not end-to-end and required manual post-search fine-tuning.
Therefore, in this work, we attempt to comprehensively study this new direction by searching for novel neural structures from a larger design space and a finer granularity without the limitation of transformation equivalence and achieve a user-friendly end-to-end system.

\section{\ideafull}
\label{sec:idea}

\begin{figure*}
	\centering
	\includegraphics[width=\linewidth]{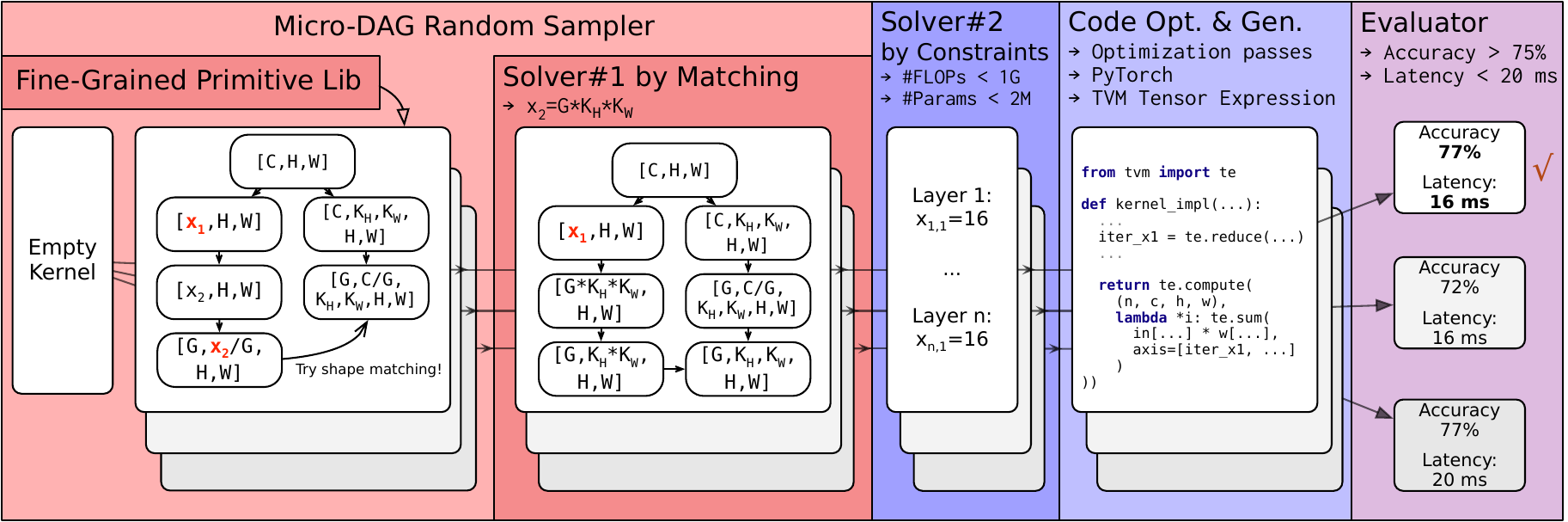}
	\caption{System overview of \name and its workflow.
	The \udag random sampler generates a kernel that contains two dynamic variables, $x_1$ and $x_2$.
	The solver by shape matching sets the value of $x_2$ when merging the two branches into one.
	The solver by constraints assigns concrete values to the remaining $x_1$ for each replacement target.
	The code generator implements the resultant kernels.
	Finally, the evaluator selects the best candidate according to accuracy and runtime measurements.
	}
	\label{fig:overview}
        \vspace{-10pt}
\end{figure*}

The similarity between tensor program optimizations and NAS techniques motivates us to take one step further along the path.
Specifically, we propose \emph{\ideafull} (\idea). \idea is a new paradigm to construct effective and efficient tensor computation kernels, the basic building blocks in NNs.
The main innovation of \idea is to \emph{use system techniques to search for novel kernel designs with high accuracy and high performance}.
We view it as neither a compiler nor a pruner/tuner but an automated explorer for new kernels.
This is because we are not transforming existing kernels but constructing new kernels from scratch.
In the long term, we aim at automatic algorithmic exploration from the system perspective under the concept of \idea.

\idea treats the computation of a kernel as a micro directed acyclic graph (\udag) $G=(V, E)$. Each node $v_i \in V$ represents the current tensor shape, and edge $e_{ij}\in E$ represents a fine-grained primitive that transforms tensor $v_i$ to the output tensor $v_j$.
Fine-grained primitives offer a lower-level representation than traditional coarse-grained kernels like convolutions and matrix/tensor multiplications.
They enable us to replace the monolithic and heavy kernel into a DAG of cheap primitives with a smaller total cost.

In \idea, we apply a new perspective to balance performance and accuracy by \emph{treating runtime performance as first-priority constraints} and searching for kernels that achieve the best accuracy within the given performance limit.
Such a ``performance-first'' philosophy essentially flips the conventional workflow, which first designs NN models with certain accuracy levels and then uses tensor compilers to optimize performance.
The new approach allows us to balance the two objectives better with a smoother flow in our system. 

\textbf{Design challenges.}
Nevertheless, realizing a practical \idea framework to automatically and efficiently explore the huge design space is still challenging.
We summarize the key questions below, which our proposed system has to address.

\begin{itemize}[topsep=0.15pt, itemsep=0.15pt, partopsep=0.15pt, parsep=0.15pt, leftmargin=8pt]
    \item What are the necessary fine-grained primitives \idea must incorporate to build high-quality kernels to allow for both rich expressibility and flexible construction?
    \item What sampling and search algorithms should \idea use to construct candidate kernels? How to balance the use of different primitives?
    \item How to effectively determine the legality of generated kernels with complex topologies, particularly when multiple branches interact and require matching dimension sizes?
    \item What interface and techniques should \idea use to satisfy various user-specified constraints, including FLOPs, parameter numbers, runtime, and accuracy?
\end{itemize}

\section{\name Design Overview}
\label{sec:overview}

As a concrete realization and an early milestone of \idea, we design \emph{\name} (CONVolution-like Architecture Search), an end-to-end, automated, and efficient framework.
\cref{fig:overview} illustrates an overview of the system components in \name and their workflow.

\begin{figure}
	\centering
	\includegraphics[width=0.7\linewidth]{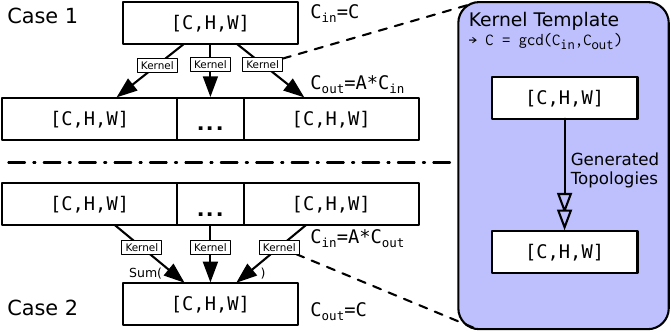}
	\caption{A single kernel template of \tensorshape{C, H, W} $\Rightarrow$ \tensorshape{C, H, W} is applied to general convolutions with different $C_{\text{in}}$ and $ C_{\text{out}}$ values.}
	\label{fig:kernel-template}
        \vspace{-13pt}
\end{figure}

\textbf{Main focus: convolutions.}
Generally, \idea can be made to generate any type of kernels and integrate them into any NN backbone topology.
In \name, we mainly focus on \emph{Convolution} Neural Networks (CNNs), which are the state-of-the-art solutions in many real-world application scenarios~\cite{liu2021swin,liu2022convnet,yuan2021incorporating}.
A convolution kernel takes a tensor of shape \tensorshape{C_{\text{in}}, H, W} as input, uses a set of $K_H \times K_W$ filters to aggregate information from neighbor pixels on multiple channels, and produces a \tensorshape{C_{\text{out}}, H, W} output tensor.
With \name, our goal is to generate new kernels with the same input and output shapes but with entirely different computational behaviors that improve performance and/or accuracy. Then we could replace standard convolutions with the new kernels in the same backbone NN topology.
For simplicity, \name only searches for \emph{a single kernel template}, which takes an input tensor of \tensorshape{C, H, W} and produces an output tensor of the same shape \tensorshape{C, H, W}.
For a general convolution of \tensorshape{C_{\text{in}}, H, W} $\Rightarrow$ \tensorshape{C_{\text{out}}, H, W}, we observe that usually one of $C_{\text{in}}$ and $C_{\text{out}}$ is a multiple of the other, so it can be composed using the kernel template as shown in \cref{fig:kernel-template}.
This allows \name to focus on optimizing a single kernel template shaped of \tensorshape{C, H, W} $\Rightarrow$ \tensorshape{C, H, W}, amortizing the expensive search cost.

\textbf{Interface and functionality.}
With a simple Python interface \codeword{canvas.sample(nn, budget)}, users specify a backbone \codeword{nn}, and designates \name to sample a new kernel under the system \codeword{budget}.
The user can specify \codeword{budget} using a variety of constraints and optimization goals, which we put into two categories. 
The first category includes constraints that can be \emph{analytically calculated}, e.g., numbers of FLOPs and model parameters. 
The other category covers the constraints and goals that require \emph{experimental measurements}, e.g., latency and accuracy.

\textbf{Workflow.}
Following \cref{fig:overview}, we briefly introduce the end-to-end workflow of \name.
\name composes a library of various \emph{fine-grained primitives} as the building blocks (\cref{sec:primitives}) and uses a \emph{\udag random sampler} to build candidate kernel implementations.
Starting from the input tensor, it extends the current \udag by randomly sampling from the primitive library (\cref{sec:kernel-gen:sampling}). The sampler can also grow the \udag into multiple branches. Finally, it merges the branches back because of the hard shape constraint of the single output tensor \tensorshape{C, H, W}\footnote{The kernel in \cref{fig:overview} is unfinished, a complete kernel after sampling should have exactly one output tensor shaped \tensorshape{C, H, W}.}.

However, some primitives (e.g., fully connected) may introduce intermediate tensors with arbitrary shapes. We use \emph{dynamic variables} (denoted as $x_i$) to represent such free dimensions and use a \emph{solver by shape matching} to coordinate the free variables in different branches so that they can match (\cref{sec:kernel-gen:variables}).
There could still be some dynamic variables left undetermined after the kernel structure is finalized. We use a \emph{solver by constraints} (\cref{sec:kernel-gen:constraints}) to solve their values according to the analytical constraints, e.g., numbers of FLOPs and parameters.

After all its dynamic variables are set, we apply each kernel candidate to the backbone NN. The code generator generates optimized code, and the evaluator uses a distributed platform to experimentally measure the rest constraints and goals.

\section{Fine-Grained Primitives}
\label{sec:primitives}

\name uses a set of fine-grained primitives as the edges of the \udag.
It is critical to designate the appropriate granularity for these primitives. 
However, neither the techniques from the current system nor machine learning research satisfies this demand.
The instruction and loop levels in tensor compilers are overly elaborative and would make the design space of constructing a kernel from scratch too large to be tackled; the existing NAS methods only work in a coarse-grained manner and offer limited flexibility to fine-tune performance and accuracy.

Instead, we observe that most of the neural operators, especially those applied on multi-dimensional tensors and aggregating neighbor information, could be decomposed into \emph{data rearrangements} and \emph{actual arithmetic computations}.
For example, a standard convolution first \emph{unfolds} the input $I$ of shape \tensorshape{C, H, W} using a receptive view \tensorshape{K_H, K_W}, and produces a tensor $U$ of shape \tensorshape{C, K_H, K_W, H, W}.
This step does not actually copy neighbor pixels but rearranges data through shallow indexing: $U(c, kh, kw, h, w) = I(c, h+kh, w+kw)$ (lowercase variables represent iterators while uppercase ones represent constants).
Then, the actual computation is done by multiplying tensor $U$ with a weight tensor to obtain the result.
This decoupling has been there since convolutions appear, known as \codeword{im2col}~\cite{chellapilla2006high}. It reflects the semantic understanding gap between system programmers and machine learning researchers.
Similar reasoning can also be found in many other operators, especially those designed for lightweight computations: group convolution uses an rearrangement like $U(g, c, kh, kw, h, w) = I(g \times \text{GROUP\_SIZE} + c, h+kh, w+kw)$; shift convolution applies an offset on a spatial dimension $U(c, h, w) = I(c, h + 1, w)$.

Following the above idea of decomposition, we conclude that the granularity of \name needs to follow this perspective of \emph{rearrangement and computation decoupling}, which is more efficient than compiler primitives and more flexible than NAS building blocks.
Moreover, to support complex topologies in a \udag, primitives are needed to blend (i.e., merge) multiple branches.
Consequently, we design our fine-grained primitive library in \name, which could be categorized into three classes summarized in \cref{tab:primitive-table}.

\begin{table*}
\centering\tablefontsize
\caption{Three classes of fine-grained primitives used in \name.}
\label{tab:primitive-table}
\begin{threeparttable}
\begin{tabular}{ccl}
\toprule
\textbf{Class} & \textbf{Primitive} & \textbf{Description and Shape Changes} \\
\midrule
\multirow{3}{*}{Rearrangement}
 & Group & \makecell[l]{Group a channel dim $X$ by a factor $G$ or make each individual channel as a group: \\ \tensorshape{\cdots, X, \cdots} $\Rightarrow$ \tensorshape{\cdots, G, \frac{X}{G}, \cdots} \text{or} \tensorshape{\cdots, X, 1, \cdots}} \\
 & Shift & Shift by 1 pixel at a spatial dim: \tensorshape{\cdots, X, \cdots} $\Rightarrow$ \tensorshape{\cdots, X_{+1}, \cdots} \\
 & Unfold & \makecell[l]{Unfold $K$ neighbors of spatial dim $X$ to a channel dim: \\\tensorshape{\cdots, X, \cdots} $\Rightarrow$ \tensorshape{\cdots, K_{X}, \cdots, X, \cdots}} \\
\midrule
\multirow{4}{*}{Arithmetic}
 & Fully-connected & Remap values at all channels to a new dim $x$: \tensorshape{\cdots_{\text{channel}}, \cdots} $\Rightarrow$ \tensorshape{x, \cdots} \\
 & Element-wise & ReLU, $\text{abs}$, $\sin$, $\exp$, etc.: \tensorshape{\cdots} $\Rightarrow$ \tensorshape{\cdots} \\
 & Folding & Average/max pooling at any dim $X$: \tensorshape{\cdots, X, \cdots} $\Rightarrow$ \tensorshape{\cdots, \cdots} \\
  & Softmax & Softmax at any continuous dims: \tensorshape{\cdots, \{\cdots\}, \cdots} $\Rightarrow$ \tensorshape{\cdots, \{\cdots\}, \cdots} \\
\midrule
\multirow{1}{*}{Blending}
 & Broadcast & Broadcasting add/sub/mul/min/max from LHS to RHS: $[\cdots]_{\text{LHS}}$, $[\cdots]_{\text{RHS}}$ $\Rightarrow$ $[\cdots]_{\text{RHS}}$ \\
\bottomrule
\end{tabular}
\end{threeparttable}
\vspace{-10pt}
\end{table*}

\textbf{Rearrangement primitives.}
The primitives in this class generalize the first type of operations in the above example.
These primitives enable flexible data rearrangement across different channels and spatial dimensions in the tensor.
For example, along a channel dimension, we could \emph{group} the data into $G$ separate groups to apply subsequent intra-group transformations.
With the \tensorshape{H, W} spatial pixels, we could \emph{shift} the pixels to manipulate neighbor information.
Finally, the \emph{unfold} primitive extracts spatial neighborhood information into the channel space.

\textbf{Arithmetic primitives.}
This primitive class includes the arithmetic operators commonly used in NNs, which change the numerical value of a tensor.
The most common one is \emph{fully-connected} (FC).
It remaps all the channel dimensions into a new dimension with an arbitrary size, denoted by a dynamic variable $x$ as the number of output channels.
In our design, FC is the only primitive that contains learned parameters, a.k.a. weights.
It is an essential primitive for NN functionality, but it also incurs higher cost than others, as well as the need to handle dynamic variables.
Besides, we have \emph{element-wise/activation} functions (ReLU, $\text{abs}$, etc.), \emph{folding} (average pooling at a certain dimension), and \emph{softmax}.
All of them are simple and cheap operators but could be helpful to introduce non-linearity and other properties to the data.

\textbf{Blending primitives.}
To eventually produce a single output tensor, the kernel \udag must merge the multiple branches which may grow during the random sampling process (\cref{sec:kernel-gen:sampling}).
Having such a capability allows us to support advanced complex connections, such as residual blocks (broadcasting addition)~\cite{resnet}.
To blend two tensors into one, we introduce the \emph{broadcast} primitive. It takes two tensors $\mathrm{LHS}$ and $\mathrm{RHS}$ as inputs and produces an output tensor with the same shape as $\mathrm{RHS}$.
To do so, we denote the two input shapes in the form of \tensorshape{\cdots_{\text{common\ prefix}}, \cdots_{\mathrm{LHS}}, \cdots_{\text{common\ suffix}}} and \tensorshape{\cdots_{\text{common\ prefix}}, \cdots_{\mathrm{RHS}}, \cdots_{\text{common\ suffix}}}. The data on the dimensions \tensorshape{\cdots_{\mathrm{LHS}}} are broadcast to \tensorshape{\cdots_{\mathrm{RHS}}}, i.e., replicated multiple times and applied to $\mathrm{RHS}$ through a binary operation like addition, subtraction, or multiplication.
This requires the total size of \tensorshape{\cdots_{\mathrm{LHS}}} dimensions must be a factor of that of \tensorshape{\cdots_{\mathrm{RHS}}}.
For example, broadcasting \tensorshape{G, \frac{x_1}{G}, H, W} to \tensorshape{G, \frac{C}{G}, K_H, H, W} expects that $\frac{x_1}{G}$ is a factor of $\frac{C}{G} \times K_H$.
Such dimension matching introduces a new challenge about dynamic variable substitution, which we resolve in \cref{sec:kernel-gen:variables}.

\textbf{Expressibility of our primitive library.}
The fine-grained primitive library is sufficiently expressive to explore a large design space of kernel construction, including complex manual designs by previous works.
Take Involution~\cite{Li_2021_CVPR} for a detailed illustration, which is embedded in \cref{fig:overview}.
The central part is a broadcast multiplication between two tensors \tensorshape{G, K, K, H, W} and \tensorshape{G, C/G, K, K, H, W}.
We can first generate them from the input tensor using group and unfold primitives, temporally as \tensorshape{G, x_2/G, H, W} and \tensorshape{G, C/G, K, K, H, W}.
Then we try to blend them with broadcasting and substitute $x_2=GKK$. 
In the end, we still have a free variable $x_1$, which can be easily scaled to realize different FLOPs and model sizes (\cref{sec:kernel-gen:constraints}).
Note that this is just one specific sample in the large design space. 
In our experiments, we indeed observe that almost all manual designs appear during the search or in the final results, among other previously unexplored constructions.

\section{Kernel Generation}
\label{sec:kernel-gen}

In \name, the \udag random sampler stochastically and iteratively generates a large number of candidate kernels by sampling from the primitive library.
\cref{sec:kernel-gen:sampling} describes the sampling algorithm and our pruning techniques.
A key challenge is to assign values to dynamic variables to ensure legality.
We first propose a variable solver in \cref{sec:kernel-gen:variables} to address the problem in the sampling process.
Finally, \cref{sec:kernel-gen:constraints} resolves any remaining variables after primitive sampling according to the analytical constraints.

\subsection{Sampling Algorithm}
\label{sec:kernel-gen:sampling}

Given the high flexibility of constructing complex \udag{}s from the rich set of primitives, \name uses random sampling.
\name builds the \udag from one node with the shape of input tensor: \tensorshape{C, H, W}. We iteratively grow it to $N$ nodes, where $N$ is a hyperparameter.
In the step $t$, we have $n_t=|V_t|$ nodes in $g_t$. We calculate all possible single-input primitives $\{p^i_t\}$ for each node $v^i_t$ and blending primitives $\{p^{i,j}_t\}$ for each pair of nodes $(v^i_t,v^j_t)$. Then one primitive $e_t$ is selected from $\Tilde{E_t}=\{\{p^1_t\},\{p^2_t\},...,\{p^{n_t}_t\},\{p^{1,2}_t\},\{p^{1,3}_t\},...,\{p^{n_t-1,n_t}_t\}\}$. Then $E_{t+1}=\{E_t,e_t\}$. After deciding the $e_t$, we add to $g_t$ the new node derived from input node(s) of $e_t$ and get $g_{t+1}$.  

\textbf{Sampling probability adjustment.}
A key innovation in the random sampler is the need to carefully adjust the sampling probability of each primitive.
There are two reasons.
First, notice that the number of choices from $\{p^i_t\}$ and $\{p^{i,j}_t\}$ are drastically different, i.e., $O(n)$ vs. $O(n^2)$. Uniformly sampling from $\Tilde{E_t}$ would result in a large bias towards the blending primitives.
We hence re-scale the probability of each primitive so that the sampling is uniform w.r.t. each primitive type, regardless of $n_t$.


\textbf{Topology heuristics.}
Moreover, \udag is constrained by certain topology properties. We must ensure that $n_T=N$ for the final step $T$, and $n_T = n_{T-1} + 1$.
Therefore, we need to heuristically restrict the number of leaf nodes in $g_t$, which we term as $W(g_t)$.
During construction, we examine the number of remaining primitives $N - n_t$ to see whether we are allowed to further extend more branches or must start to merge.
For example, if $W(g_t)=3$ and $N - n_t=2$, we must combine two tensors.
We control such behaviors by modifying the sampling probability of primitives, e.g., by setting all probabilities to 0 except for blending primitives.

\textbf{Pruning techniques.}
Random sampling is an expensive process. We apply several pruning heuristics to improve its efficiency.
We use an approximate graph isomorphism hash function similar to~\cite{douglas2011weisfeiler} and deduplicate generated kernels.
Besides, we eliminate obviously sub-optimal constructions with redundant or illegal components, such as consecutive ReLUs and subtracting two equal tensors.
With such pruning, \name is able to sample a legal kernel candidate in the huge search space within a few milliseconds.

\subsection{Dynamic Variable Solver by Shape Matching}
\label{sec:kernel-gen:variables}
During the sampling process mentioned above, the FC primitives in our candidate \udag produce free dynamic variables.
These dynamic variables further enlarge the search space and provide more flexibility to tune the numbers of FLOPs and parameters in our kernel for performance-accuracy tradeoffs.
Recall that an FC primitive uses weights to remap \tensorshape{\cdots_{\text{channel}}, \cdots_{\text{spatial}}} into \tensorshape{x, \cdots_{\text{spatial}}}, where $x$ indicates that the output channel dimension can have any size.
Other primitives typically have deterministic dimensions without introducing variables.

The concrete values of all dynamic variables will be eventually assigned according to the constraints such as FLOPs and model sizes (\cref{sec:kernel-gen:constraints}).
However, not all variables can be independently set in the \udag.
In particular, the blending primitive that combines two tensors requires matched dimension sizes of the two inputs.
For example, when a tensor \tensorshape{x_1, H, W} is broadcast to another tensor \tensorshape{C, K_H, H, W}, $x_1$ must be a factor of $CK_H$.
By default, the group number $G$ is a factor of $C$. So $x_1$ can only take a value from the set $\{C, K_H, G, \frac{C}{G}, CK_H, GK_H, \frac{C}{G}\times K_H\}$, which is an additional constraint that must be satisfied.

To handle such constraints among dynamic variables in a general way, we first prove a simple theorem.

\begin{theorem}\label{thm:dyn-var}
For any tensor constructed in \name where all its dimensions are in the form of $D=\frac{\text{numerator}}{\text{denominator}}$ (e.g. $D=\frac{C}{G}$ or $D=CK_H$), there is
\begin{itemize}[topsep=0.15pt,itemsep=0.15pt,partopsep=0.15pt, parsep=0.15pt, leftmargin=8pt]
    \item no dynamic variable in any spatial dimension (i.e., \tensorshape{C, H, x_0} does not exist);
    \item no dynamic variable in the denominator of a channel dimension (i.e., $\frac{C}{x_0}$ does not exist);
    \item at most one dynamic variable in the numerator of a channel dimension (i.e., $\frac{x_0x_1}{K_H}$ does not exist);
    \item at most one dynamic variable across all dimensions (i.e., \tensorshape{x_0, x_1, H, W} does not exist).
\end{itemize}
\end{theorem}

\textbf{Proof sketch.}
We prove this by induction. For the input tensor shaped \tensorshape{C, H, W}, the theorem naturally holds.
For all primitive types except FC, the shape transformation neither introduces new dynamic variables nor moves an existing dynamic variable to a denominator.
For an FC primitive, all channel dimensions are remapped to a new dimension with a single dynamic variable $x$.
With the fact that no primitives introduce dynamic variables into spatial dimensions, the theorem holds.
\hfill$\square$

\cref{thm:dyn-var} simplifies the design to substitute dynamic variables for broadcast primitives in \name.
For two tensors' shape denoted as \tensorshape{\cdots_{\text{common\ prefix}}, \cdots_{\mathrm{LHS}}, \cdots_{\text{common\ suffix}}} and \tensorshape{\cdots_{\text{common\ prefix}}, \cdots_{\mathrm{RHS}}, \cdots_{\text{common\ suffix}}},
we first detect and strip the common prefix and suffix and focus on the unmatched parts $\{\cdots_{\mathrm{LHS}}\}$ and $\{\cdots_{\mathrm{RHS}}\}$.
We calculate their total sizes as $\#\{\cdots_\mathrm{LHS}\}=\frac{\Pi_i \text{numerator}_{\mathrm{LHS}, i}}{\Pi_i \text{denominator}_{\mathrm{LHS}, i}}$, and similarly for $\#\{\cdots_\mathrm{RHS}\}$.
\cref{thm:dyn-var} says that only one dynamic variable could exist in the numerator of each, which we use $k_{\{\mathrm{LHS},\mathrm{RHS}\}} \in \{0, 1\}$ to indicate.
So if we take the ratio $M$ between the two sizes, we have:
\begin{equation*}
M=\frac{\#\{\cdots_\mathrm{RHS}\}}{\#\{\cdots_\mathrm{LHS}\}}
=\frac{
(x_{\mathrm{RHS}})^{k_\mathrm{RHS}}R_{\mathrm{RHS}}
}{
(x_{\mathrm{LHS}})^{k_\mathrm{LHS}}R_{\mathrm{LHS}}
}
\end{equation*}
where $R_{\mathrm{LHS}}$ and $R_{\mathrm{RHS}}$ are fully simplified as remaining factors.

To ensure legality in broadcast, $M$ must be an integer, which determines the possible values for dynamic variables $x_{\mathrm{LHS}}$ and $x_{\mathrm{RHS}}$ (if existing).
This mainly constrains $x_{\mathrm{RHS}}$ in the denominator, which should be a factor of the numerator.
Naturally, the valid substitutions of $x_{\mathrm{LHS}}$ are the set of all factors of $(x_{\mathrm{RHS}})^{k_\mathrm{RHS}}R_{\mathrm{RHS}}$.
As an example, assume two tensors \tensorshape{x_1, K_H, H, W} and \tensorshape{G, \frac{x_2}{G}, K_H, K_W, H, W}.
The common prefix is empty, and the common suffix is $\{H, W\}$, which are stripped.
The substitutions of $x_1$ could be all the factors of $x_2 K_W$, i.e., $\{1, x_2, K_W, x_2K_W\}$.

Once we have the valid substitution set from the dynamic variable solver, the random sampler will randomly select one value and propagate to the whole \udag.
For example, assume we sample $x_1 = x_2 K_W$, then the shape of $\mathrm{LHS}$ becomes \tensorshape{x_2 K_W, K_H, H, W}.

\subsection{Dynamic Variable Solver by Analytical Constraints}
\label{sec:kernel-gen:constraints}

After some of the dynamic variables are solved in the generation process, the remaining variables should still be carefully considered and assigned according to all \emph{analytical constraints}, e.g., FLOPs and parameter sizes.

Generally speaking, the generated kernel template needs to substitute multiple standard convolutions in an NN.
Each individual convolution has a specific assignment of $C$, $K_H$, $K_W$, $H$, $W$, $G$, as well as the dynamic variables $x$.
For these remaining dynamic variables, we denote them as $x_{i,j}$, i.e., the $j$-th variable in the $i$-th convolution target.
The constraint solver derives all values of $x_{i,j}$ using a heuristic algorithm to maximally utilize the available budget of the constraints that can be analytically modeled, e.g., FLOPs and parameter sizes.

Before dealing with $x_{i,j}$, we first specially handle the group number $G$.
We observe that most of the modern NN designs that adopt group convolutions~\cite{resnext, regnet} usually apply the same number of groups in almost all their layers.
Therefore we enforce a global value $G$ for all the group primitives across \emph{all} targets in the NN.
$G$ must be a common factor of the original channel numbers $C_i$ of these replaced convolutions, i.e., $G$ is a factor of $\gcd(C_i)$.
For example, with two targets of $C_1=32$ and $C_2=48$, $G$ should be sampled from a set of $\{2, 4, 8, 16\}$.
It simplifies our dynamic variable solver by excluding $G$ from free variables.

Now, all remaining $x_{i,j}$ variables were generated by FC primitives as the output channel dimensions. They directly contribute to the FLOPs, the number of parameters, and model accuracy.
Our heuristic algorithm sets their values in two steps.
First, as each $x_{i,j}$ denotes a channel dimension, it should roughly match the channel number of the input/output tensor at the target. The solver thus derives the base (i.e., minimum) value of each $x_{i,j}$ that satisfies structural legality and is proportional to the input channel number $C_i$.
Second, more channels (thus higher FLOPs and more parameters) usually help improve model accuracy.
So the solver tries to fully utilize the allowed constraint budget by using the maximum value for each $x_{i,j}$, which is a multiple of the above base $x_{i,j}$.

Specifically, we should first satisfy the legality of individual dimension sizes and broadcast primitives, similar to \cref{sec:kernel-gen:variables}. 
Recall from \cref{thm:dyn-var} that any $x_{i,j}$ only appears in numerators, i.e., the dimension size is in the form of $\frac{p x_{i,j}}{q}$.
Each variable may appear in more than one primitive in each kernel; so for each $x_{i,j}$, there may exist multiple pairs of $(p, q)$ (after fully reduced, $\gcd(p, q)=1$). To ensure integer dimension sizes, we must have:
\begin{equation*}
x_{i,j} = k_{i,j} \times \mathrm{lcm}(q_1, q_2, \cdots) \overset{\text{def}}{=} k_{i,j} \times \mathrm{lcm}_{i,j}, \quad k_{i,j} \in \mathbb{Z}^+
\end{equation*}
where $\mathrm{lcm}_{i,j}$ is the least common multiple of all $q$s.

To make each $x_{i,j}$ proportional to the corresponding $C_i$, without loss of generality, we assume $C_1 = \min_i(C_i)$. We require approximately that:
\begin{equation*}
\frac{x_{i,j}}{x_{1,j}}
=\frac{k_{i, j} \times \mathrm{lcm}_{i,j}}{k_{1, j} \times \mathrm{lcm}_{1,j}}
\approx\frac{C_i}{C_1}
\end{equation*}
By setting $k_{1, j}=1$, we obtain a series of the base $x_{i,j}$ values that satisfy legality and heuristically retain channel scaling throughout the network:
\begin{equation*}
k_{i,j} = \frac{C_i \times \mathrm{lcm}_{1,j}}{C_1\mathrm{lcm}_{i,j}}
    \approx \lceil \frac{C_i \times \mathrm{lcm}_{1, j}}{C_1\mathrm{lcm}_{i,j}} \rceil,
\, x_{i,j} = k_{i,j} \times \mathrm{lcm}_{i,j}
\end{equation*}
In the second step, we fully utilize the budget of each analytical constraint.
Assume a constraint $\text{Cstr}(\text{net}(\{x_{i,j}\})) \leq \text{budget}$, where $\text{Cstr}$ could be FLOPs, parameter sizes, or other constraint functions.
If even the previously solved base $x_{i,j}$ values violate the constraint, we discard this kernel and sample the next.
Otherwise, we try to increase these variables according to their utilization sensitivities. Specifically, we calculate each $\Delta_{i,j}=\text{Cstr}(\text{net}(\cdots, 2x_{i, j}, \cdots)) - \text{Cstr}(\text{net}(\cdots, x_{i, j}, \cdots))$, and double the $x_{i,j}$ by the ascending order of $\Delta_{i,j}$ at each iteration, until we cannot further increase any $x_{i,j}$ without exceeding the budget.

\begin{figure}
	\centering
	\includegraphics[width=0.65\linewidth]{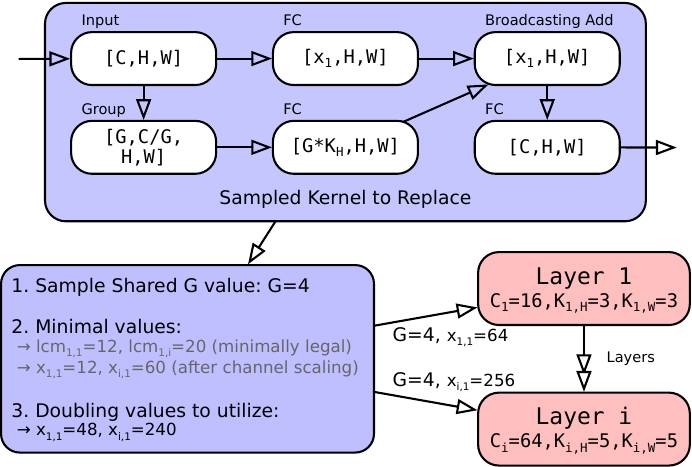}
	\caption{An example of solving dynamic variables for different convolution targets by the constraint solver.}
	\label{fig:solve-dynamic-var}
        \vspace{-15pt}
\end{figure}

\cref{fig:solve-dynamic-var} illustrates an example, where the broadcast primitive between \tensorshape{GK_H, H, W} and \tensorshape{x_1, H, W} requires $x_1$ must be a multiple of $G K_H$. 
Applying it to the two layers gives $\mathrm{lcm}_{1, 1}=GK_{1, H}=12$ and $\mathrm{lcm}_{i, 1}=GK_{i, H}=20$. 
Scaling by $C$, we have $x_{1, 1}=12$ and $x_{i, 1}=\lceil\frac{C_i\mathrm{lcm}_{1, 1}}{C_1\mathrm{lcm}_{i, 1}} \rceil x_{1, 1}=60$.
Finally, to fully utilize the budget, we double the values two times and get $x_{1, 1}=48$ and $x_{i, 1}=240$.

\section{Kernel Evaluation}
\label{sec:kernel-eval}

After substituting all variables, we now have a concrete kernel instance for each replacement target.
\name next generates code implementation and measures the runtime and accuracy on real hardware platforms.

\textbf{Code generation and runtime evaluation.}
\name does code generation for PyTorch and TVM separately.
It translates the kernels to \codeword{nn.Module} classes in PyTorch and uses them to build the entire model for training.
We also translate kernels into TVM Tensor Expression language~\cite{autotvm} and use TVM Ansor~\cite{zheng2020ansor} for performance-oriented machine code generation and tuning.
The code generation also includes several optimizations passes, e.g., adding normalization for numeric stability and other common compiler optimizations.

\textbf{Model accuracy evaluation.}
Fully training a typical NN usually takes hours to days, depending on the model size.
Existing NAS techniques reduce such overheads by using proxy datasets~\cite{zoph2018learning}, fewer epochs~\cite{zoph2018learning, ha2016hypernetworks}, zero-cost estimation~\cite{abdelfattah2021zero, baker2017accelerating}, or other early-pruning techniques.
All these solutions could be directly incorporated in \name.
We also developed a new pruning strategy that we empirically find very effective.
We record the accuracy curve (accuracy vs. epochs) of the best accuracy result achieved so far.
When training a new candidate, we require the accuracy at each epoch to be at least a given fraction of the accuracy of the recorded best result at the same epoch, i.e., $\text{accuracy}_{\text{ker}}(\text{epoch}) \geq \lambda \cdot \text{accuracy}_{\max}(\text{epoch})$, where $\lambda = f(\text{epoch})$ and $f(x)=\theta+(1-\theta)x$. Basically, $\lambda$ is small at early epochs and gets larger later, so we allow a kernel to perform less effectively at an early stage but eventually should approach the best result.
$\theta$ adjusts the pruning strictness and is typically set to 0.5.

\textbf{Implementation: distributed evaluation infrastructure.}
We develop a platform to automatically dispatch runtime/accuracy evaluation tasks to a distributed pool of workers, e.g., GPUs for accuracy tasks and CPU cores for runtime tasks.
Due to the high parallelism in our random sampling algorithm and the complete independence of all tasks, this infrastructure has superior scalability. This is desired as it offers a nice benefit vs. cost tradeoff: the more computing resources you use, the higher rewards in terms of better kernel implementations you may get.
In contrast, more advanced algorithms such as evolutionary search~\cite{liu2021survey} may suffer from the inefficiency caused by heavy task dependency and cannot scale well.
We implemented all the procedures as a distributed and end-to-end framework with 8,000 lines of C++ and 6,000 lines of Python.

\section{Results}
\label{sec:evaluation}

\subsection{Experimental Setups}

\noindent\textbf{Hardware configurations.}
We use a cluster of four nodes, each with two sockets of 20-core Intel\textregistered\ Xeon\textregistered\ Gold 5218R processors, 256 GB of DRAM, and four RTX\texttrademark\ 3090 GPUs.

\noindent\textbf{Workloads.}
We choose 8 commonly used NNs: ResNet-18~\cite{resnet}, ResNet-29~\cite{resnet}, VGG-16~\cite{simonyan2014very}, DenseNet-161~\cite{huang2017densely}, MobileNet-V2~\cite{sandler2018mobilenetv2}, ResNeXt-29-2x64D~\cite{resnext}, RegNet-X-800MF~\cite{regnet}, and MNASNet-1-0~\cite{tan2019mnasnet}.
We set the replacement targets to be all standard convolutions in these NNs.

\noindent\textbf{Baselines.}
We use two baselines for comparison.
TVM Ansor~\cite{zheng2020ansor} is a state-of-the-art tensor compiler that preserves mathematical equivalence during optimizations and provides performance with the original models.
Turner et al.~\cite{turner} (labeled as NAS-PTE, as mentioned in \cref{sec:background}) is the first work that introduced NAS-style transformations into tensor program optimizations, but at a preliminary stage which only involves simple loop range number changing and is still in the traditional scope of convolution semantics.

\noindent\textbf{Datasets and training configurations.}
ImageNet~\cite{deng2009imagenet} has been the standard benchmark for CNNs, but it is not suitable for directly searching because of the large size.
We use the smaller but still relatively challenging CIFAR-100~\cite{cifar100} as the proxy dataset.
Specifically, NN models are fully trained for 300 epochs on CIFAR-100 (may early stop due to the pruning in \cref{sec:kernel-eval}) using stochastic gradient descent (SGD)~\cite{robbins1951stochastic}.
The selected best kernels under CIFAR-100 are then fully trained on the ImageNet dataset for 90 epochs, also using SGD, for accuracy and performance evaluation.
We scale the CIFAR-100 images to the same size as ImageNet to ensure the same inference performance.

\subsection{End-to-End Performance}

\begin{figure}
	\centering
	\includegraphics[width=0.65\linewidth]{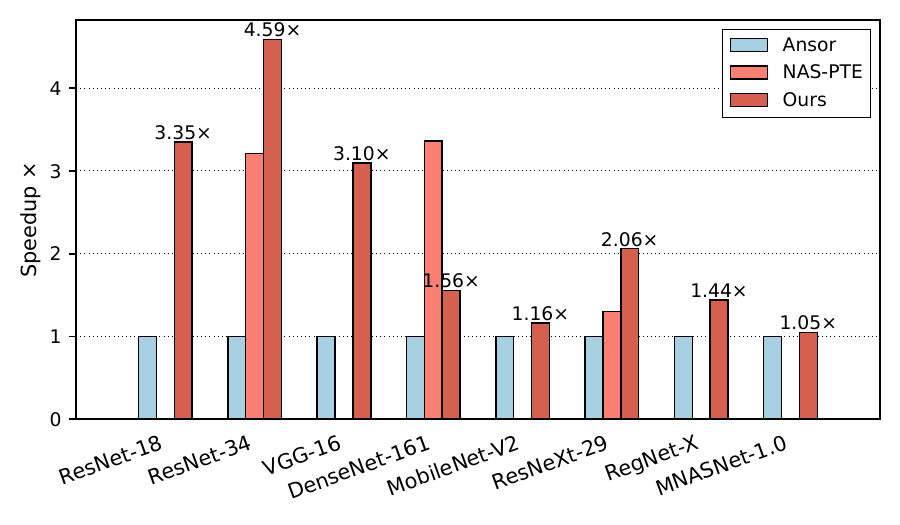}
        \vspace{-5pt}
	\caption{End-to-end performance comparison between \name and the baselines.
	}
	\label{fig:end-to-end-perf}
        \vspace{-10pt}
\end{figure}

We first compare the end-to-end performance between \name and the two baselines.
\cref{fig:end-to-end-perf} shows the performance results among all the workloads.
In \name, we keep reducing the specified FLOPs budget and see how far we can go in terms of actual runtime performance with no more than 1\% loss in CIFAR-100 accuracy.
NAS-PTE only searched for network-independent efficient kernel designs and later applied manual layer-wise optimizations to put them into an NN.
Due to the manual optimizations and code unavailability of NAS-PTE, we only report the results which can be derived from their published search results; the other NNs with missing numbers in \cref{fig:end-to-end-perf} are not evaluated.

We observe that \name achieves up to $4.6\times$ of speedup and obtain an average (geomean) speedup of $1.5\times$ across all workloads compared to the Ansor-compiled baselines.
For early NNs like ResNet-18, ResNet-34, and VGG-16, \name performs $3\times$ better due to the discovered novel kernels.
Even for relatively new and highly optimized NNs, \name is still able to achieve $1.05 \sim 2\times$ gains.
Compared to NAS-PTE, we also achieve $1.4\times$ and $1.6\times$ speedups on ResNet-34 and ResNeXt-29, even considering that NAS-PTE incorporated manual analysis and tuning and \name is fully automated.
DenseNet-161 uses a large number of convolutions with the irregular matching of input and output channel numbers, which does not satisfy \name's assumption that one is a multiple of the other. So we can only replace $60\%$ of the FLOPs that bounds the ideal speedup at $2.5\times$, while we achieve $1.56\times$.

All \name-discovered NNs have less than 1\% accuracy loss on CIFAR-100. When retrained on ImageNet, they exhibit approximately 4\% accuracy loss, which is acceptable and comparable to state-of-the-art numeric pruning techniques summarized in \cite{li2022revisiting} (reducing $\ge 50\%$ FLOPs or parameters) from the machine learning community.

\textbf{Search efficiency.}
With the efficient sampler, solver, and pruning designs, \name, as a \idea implementation, has reasonable search speedup.
During two weeks of experiments on our cluster of 16 GPUs, over 300,000 kernels were evaluated, translating to 0.01 GPU hours per kernel.

\subsection{Model Size Reduction}

\begin{figure}
	\centering
	\includegraphics[width=0.65\linewidth]{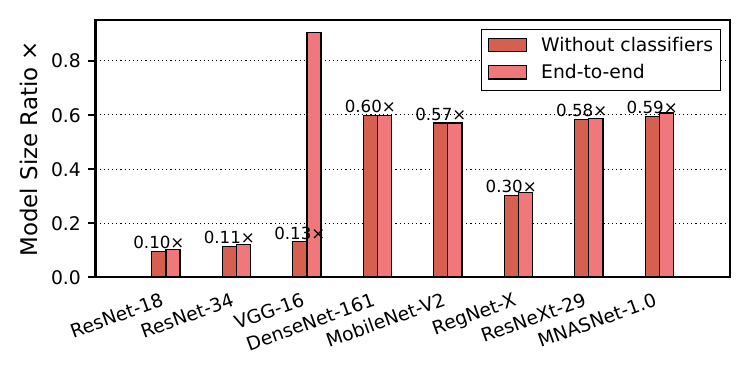}
        \vspace{-5pt}
	\caption{Model size comparison between the original size and that discovered by \name, for both the entire network and the feature extraction front-end excluding the final classifiers.
	Each number indicates the ratio of the new model size compared to the original size.}
	\label{fig:compression}
        \vspace{-5pt}
\end{figure}

We also present the model size comparison in terms of parameter numbers in \cref{fig:compression}.
Almost all models benefit from the \name design, with the new model being only $0.1\times$ to $0.6\times$ of the original size.
VGG-16 contains a significant amount ($89\%$) of model parameters in its final classifier layer. As \name only focuses on convolution replacements, we are not able to reduce the final classifier layer.
Excluding the classifier brings the model size ratio with the replaced kernel to $0.13\times$.
For classical NN designs like ResNet-18, ResNet-34, or VGG-16, we are able to compress the model size to $\sim 0.1\times$.
Even for the relatively new models, which are specifically targeted for small model sizes \emph{by design}, \name can further reduce the sizes into at least $0.5\times$.

\subsection{Kernel-Level Analysis of ResNet-34}

\begin{figure}
        \vspace{-5pt}
	\centering
	\includegraphics[width=0.65\linewidth]{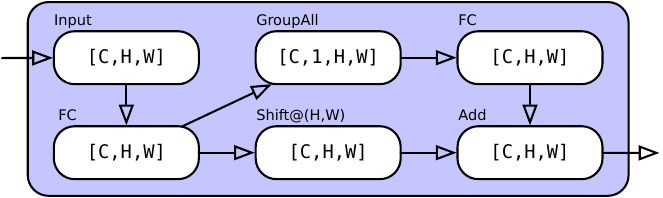}
	\caption{Best kernel found for ResNet-34.}
	\label{fig:resnet-34-best}
        \vspace{-19pt}
\end{figure}

\cref{fig:resnet-34-best} shows the best result of our search on ResNet-34.
Instead of collecting information from all neighbors by unfolding, this kernel uses a combination of shift and addition to extract features from only one adjacent pixel in order to significantly reduce FLOPs and parameters.
There is also another branch that uses a lightweight depth-wise convolution (FC within each channel separately, i.e., grouping and FC) to maintain expressive power and generalization.

\begin{figure*}
	\centering
        \vspace{-5pt}
	\includegraphics[width=\linewidth]{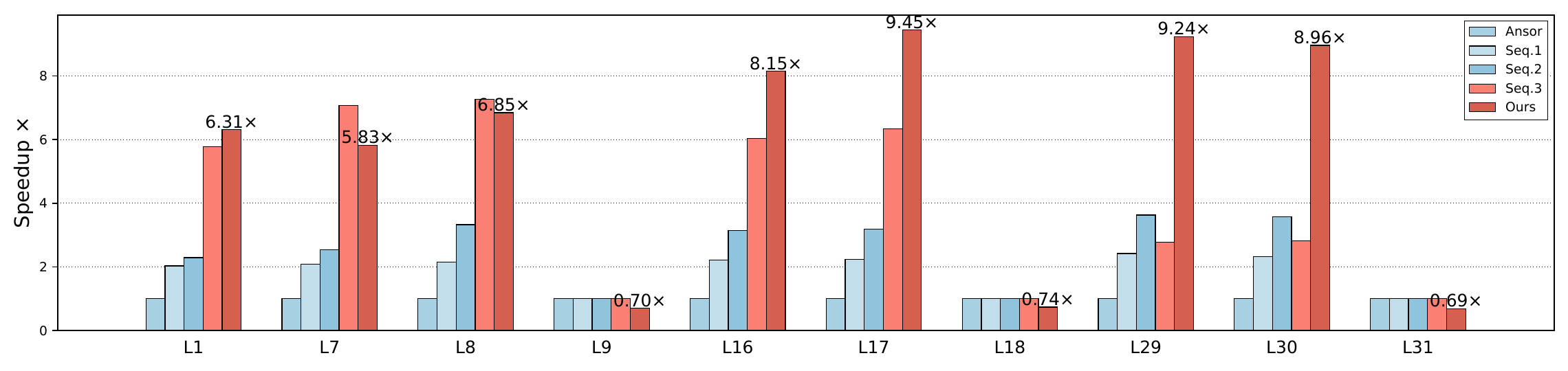}
        \vspace{-10pt}
	\caption{ResNet-34 layer-wise performance comparison between TVM Ansor, NAS-PTE, and \name.}
        \vspace{-10pt}
	\label{fig:kernel-perf}
\end{figure*}

\cref{fig:kernel-perf} summarizes all individual layer performance on ResNet-34, compared to TVM Ansor and all the three kernel structures discovered by NAS-PTE.
Our kernel outperforms the traditional convolution by up to $9.45\times$.
Compared to the optimized kernels from NAS-PTE, \name is on average $2.2\times$, $1.7\times$, and $1.2\times$ faster, respectively. These improvements are achieved without any layer-wise tuning by a fully automated workflow in \name.

\subsection{Case Study: Machine Learning Kernels}
\label{subsec:case-studies}

To understand what kernels \name can find, we study its discoveries from both historical and future views.

\textbf{Rediscovering designs in the past.}
Interestingly, \name can rediscover several convolution variants that researchers proposed previously.
For example, on VGG~\cite{simonyan2014very}, a very early linear network, the top kernels discovered by \name resemble the residual connections~\cite{resnet}, using \name's broadcast primitives.
On ResNet itself, \name finds depth-wise separable ~\cite{howard2017mobilenets}, and spatial separable ~\cite{szegedy2016rethinking} convolutions.
Finally, for MobileNet~\cite{howard2017mobilenets}, the resultant kernels include optimizations of pooling and folding, very similar to Squeeze-and-Excitation convolutions~\cite{hu2018squeeze} and Involution~\cite{Li_2021_CVPR}.

\begin{figure}
	\centering
        \vspace{-5pt}
	\includegraphics[width=0.65\linewidth]{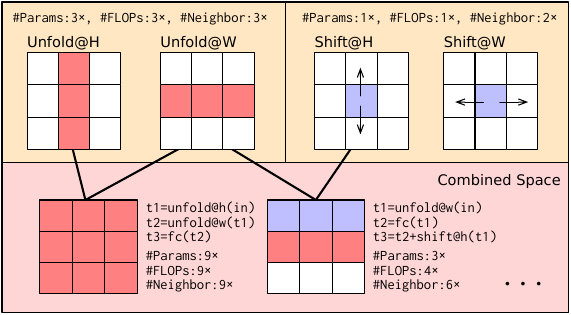}
	\caption{Fine-grained combination of unfold and shift.}
        \vspace{-10pt}
	\label{fig:primitive-combination}
\end{figure}

\name not only finds these designs but also cleverly combines them to best exploit their properties.
For example, \name expresses the aggregation of neighborhood information in one dimension and combines it with a shift primitive in another dimension to achieve lightweight but comprehensive feature extraction.
In \cref{fig:primitive-combination}, each unfold primitive can extract the information of 3 surrounding pixels. Applying unfold on both dimensions (bottom left) covers all 9 neighboring pixels, with 9 parameters and 9 FLOPs.
Instead, if we use an unfold and a shifted residual connection on the two dimensions, respectively (bottom right), we can use information from 6 neighboring pixels but reduce it to only 3 parameters and 4 FLOPs.

\begin{figure}
    \centering
\includegraphics[width=0.65\textwidth]{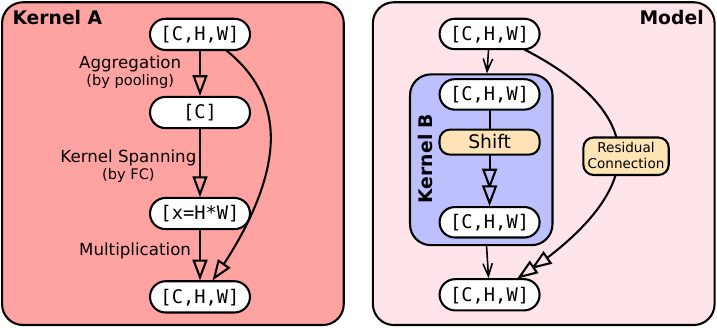}
    \caption{Two new kernel patterns discovered by \name. The first shows the dynamic variable system enables irregular designs; the second shows an out-of-kernel opportunity.}
    \label{fig:case-studies}
    \vspace{-12pt}
\end{figure}

\textbf{Motivating the future.}
\name finds many interesting and previously unexplored kernel designs with high accuracy and performance.
%
On ResNet, \name finds a design with spatial aggregation and kernel spanning, as kernel A in \cref{fig:case-studies}.
Traditional convolutions only operate on neighbor pixels.
Kernel A first aggregates \tensorshape{C, H, W} into an overall spatial vector \tensorshape{C} by pooling and then generates an FC primitive with shape \tensorshape{H \times W}. It finally multiplies this tensor with the input to get the output, which is unexpectedly similar to the popular attention mechanism~\cite{vaswani2017attention}.
With \name's variable system and shape-matching rules, such irregularities are extremely abundant in generated kernels.

A surprising finding is that, although we search for individual kernels, we end up with some primitives that do not materially affect the current kernel but are an integral part of the overall design.
For example, in kernel B in \cref{fig:case-studies}, all primitives inside the kernel only operate on channel dimensions; any spatial shift does not change the numerical results but only rearranges the pixel positions.
However, when this kernel is placed in a residual connection, the shift primitive contributes to the output of the whole residual connection.
This is a good demonstration of \name's ability to search for customized kernels in an end-to-end manner.

\section{Conclusions}

We make a case for a new paradigm of \ideafull, which stochastically explores new kernel constructions.
To demonstrate such potential, we further build an end-to-end system, \name.
\name has a random sampler to construct kernel structures from a library of fine-grained primitives, two solvers to address tensor dimension flexibility, and an evaluation system to handle analytical and experimental constraints.
Our results show \name achieves on average $1.5\times$ and up to $4.6\times$ performance improvements.

\bibliographystyle{unsrt}
\bibliography{references}

\end{document}